\documentclass[a4paper,10pt,twoside]{article}

\usepackage{clin}        
\usepackage{harvard}     

\usepackage{times}
\usepackage{latexsym}

\usepackage{url}
\usepackage[normalem]{ulem}
\usepackage{todonotes}
\usepackage{booktabs}
\usepackage{enumitem} 
\usepackage{multicol}
\usepackage{array}
\usepackage{amssymb}
\usepackage{multirow}
\usepackage[T1]{fontenc}
\usepackage{color, colortbl} 
\usepackage{soul} 

\begin{document}

\title{Unsupervised Context-Sensitive Spelling Correction of English and Dutch Clinical Free-Text with Word and Character N-Gram Embeddings}

\author{Pieter Fivez \email{Pieter.Fivez@uantwerpen.be}\\
{\normalsize \bf Simon \v{S}uster} \email{Simon.Suster@uantwerpen.be}\\
{\normalsize \bf Walter Daelemans} \email{Walter.Daelemans@uantwerpen.be}
\AND \addr{CLiPS,  University of Antwerp, Prinsstraat 13, 2000 Antwerp, Belgium} }

\maketitle\thispagestyle{empty} 


\begin{abstract}
We present an unsupervised context-sensitive spelling correction method for clinical free-text 
that uses word and character n-gram embeddings. Our method generates misspelling replacement candidates and ranks them 
according to their semantic fit, by calculating a weighted cosine similarity between the vectorized representation of a candidate
and the misspelling context. To tune the parameters of this model, we generate self-induced spelling error corpora. We perform our experiments for two languages. For English, we greatly outperform off-the-shelf spelling correction tools on a manually annotated MIMIC-III test set,
and counter the frequency bias of a noisy channel model, showing that neural embeddings can be successfully exploited to improve upon the state-of-the-art. For Dutch, we also outperform an off-the-shelf spelling correction tool on manually annotated clinical records from the Antwerp University Hospital, but can offer no empirical evidence that our method counters the frequency bias of a noisy channel model in this case as well. However, both our context-sensitive model and our implementation of the noisy channel model obtain high scores on the test set, establishing a state-of-the-art for Dutch clinical spelling correction with the noisy channel model.\footnote{Source code, which includes a script to extract the annotated English test data from MIMIC-III (for those who have access to the corpus), can be found at \url{https://github.com/clips/clinspell}. Due to privacy concerns, we are not allowed to share the annotated Dutch test data.}
\end{abstract}

\section{Introduction}

The problem of automated spelling correction has a long history, dating back to the late 1950s.\footnote{A good overview is given by Mitton (2010) and Jurafsky and Martin (2016).} Traditionally, spelling errors are divided into two categories: non-word misspellings, the most prevalent type of misspellings, where the error leads to a nonexistent word, and real-word misspellings, where the error leads to an existing word, either caused by a typo (e.g. \textit{I \sout{hole} $\rightarrow$ hope so}), or as a result of grammatical (e.g. \textit{their - there}) or lexical (e.g. \textit{aisle - isle}) confusion. The spelling correction task can be divided into three subtasks: detection of misspellings, generation of replacement candidates, and ranking of these candidate replacements to correct the misspelling. The nature of the detection subtask is dependent on the type of error: non-word misspellings are typically defined as tokens absent from a reference lexicon, while for real-word misspellings, the detection task is postponed by considering all tokens as replaceable, using the confidence of the candidate ranking module to determine which tokens should be treated as misspellings. The generation of replacement candidates is typically performed by including all items from a lexicon which fall within a pre-defined edit distance of the misspelling (e.g. all items within a Levenshtein distance of 3). The ranking component is the most complex of the three subtasks, and is the main topic of this paper.

The genre of clinical free-text poses an interesting challenge to the spelling correction task, since it is notoriously noisy. English corpora contain observed spelling error rates which range from 0.1\% \cite{liu} and 0.4\% \cite{lai} to 4\% and 7\% \cite{tolentino}, and even 10\% \cite{ruch}. Moreover, clinical text also has variable lexical characteristics, caused by a broad range of domain- and subdomain-specific terminology and language conventions. These properties of clinical text can render traditional spell checkers ineffective \cite{patrick}. Recently, Lai et al. (2015) have achieved nearly 80\% correction accuracy on a test set of clinical notes with their noisy channel model. However, their ranking model does not use any contextual information, while the context of a misspelling can provide important clues for the spelling correction process, for instance to counter the frequency bias of a context-insensitive system based on corpus frequency. As an example, consider the misspelling \textit{\sout{goint} $\rightarrow$ going} present in the MIMIC-III \cite{johnson} clinical corpus. While in many domains, \textit{going} will be a relatively frequent word type and will consequently be picked by a corpus frequency-based system, it is actually outnumbered in MIMIC-III by the more prevalent word types \textit{joint} and \textit{point}, which are other replacement candidates for the same misspelling. In other words, corpus frequency is not a reliable metric in such cases. Flor (2012) also pointed out that ignoring contextual clues harms performance where a specific misspelling maps to different corrections in different contexts, e.g. \textit{\uline{iron} \uline{deficiency} due to \sout{enemia} $\rightarrow$ anemia} vs. \textit{\uline{fluid} \uline{injected} with \sout{enemia} $\rightarrow$ enema}. A noisy channel model like the one by Lai et al. (2015) will choose the same item for both corrections.

Our proposed unsupervised context-sensitive method exploits contextual clues by using neural embeddings to rank misspelling replacement candidates according to their semantic fit in the misspelling context. Neural embeddings have proven useful for a variety of related tasks, such as unsupervised normalization \cite{sridhar} and reducing the candidate search space for spelling correction \cite{pande}. We hypothesize that, by using neural embeddings, our method can counter the frequency bias of a noisy channel model. We test our system on manually annotated misspellings from the MIMIC-III corpus. We also conduct experiments on Dutch data, since there is still a need for a Dutch spelling correction method for clinical free-text \cite{cornet}. By replicating our English research setup for Dutch, we simultaneously examine the language adaptability of our context-sensitive model, and establish a state-of-the-art for Dutch clinical spelling correction. We test our Dutch model on manually annotated misspellings from clinical records collected at the Antwerp University Hospital (UZA). In our experiments for both English and Dutch, we focus on already detected non-word misspellings for developing and testing our spelling correction method, following Lai et al (2015). Note that our method could also be applied to real-word errors. However, since our strategy for collecting an empirical test set of misspellings, which we describe in section \ref{testcorp}, can not be used for real-word errors, we do not address them in this article.

\section{Approach}

Since we focus on already detected non-word misspellings, our system only deals with two subtasks of the spelling correction task, namely, generating candidate replacements and ranking them.

\subsection{Candidate Generation}

We generate replacement candidates in 2 phases, using the reference lexicons described in section \ref{lexicon}. First, we extract all items within a Damerau-Levenshtein edit distance of 2 from a reference lexicon. Secondly, to allow for candidates beyond that edit distance, we also apply the Double Metaphone matching popularized by the open source spell checker Aspell\footnote{\url{http://aspell.net/metaphone/}}. This algorithm converts lexical forms to an approximate phonetic consonant skeleton, and matches all Double Metaphone representations within a Damerau-Levenshtein edit distance of 1. The Double Metaphone representation is an intentionally approximate phonetic representation, which is created with an elaborate set of rules, and whose principles of design include mapping voiced/unvoiced consonant pairs to the same encoding, encoding any initial vowel with `A', and disregarding all non-initial vowel sounds. For example, the Double Metaphone representation of \textit{antibiotic} is \textit{ANTPTK}.

\subsection{Candidate Ranking}

Our approach computes the cosine similarity between the vector representation of a candidate and the composed vector representations of the misspelling context, weights this score with other parameters, and uses it as the ranking criterium. This setup is similar to the contextual similarity score by Kilicoglu et al. (2015), which proved unsuccessful in their experiments. However, their experiments were preliminary. They used a limited context window of 2 tokens, could not account for candidates which are not observed in the training data, and did not investigate whether a bigger training corpus would lead to vector representations which scale better to the complexity of the task.

We undertake a more thorough examination of the applicability of neural embeddings to the spelling correction task. To tune the parameters of our context-sensitive spelling correction model in an unsupervised way, we automatically generate development corpora with artificial, randomly created spelling errors for three different scenarios following the procedures described in section \ref{devcorp}. These three types of generated spelling error corpora, which we refer to as \textit{setups}, are increasingly difficult for the spelling correction task. We apply the same setups to both English and Dutch. \textbf{Setup 1} is generated from the same corpus which is used to train the neural embeddings, and exclusively contains corrections which are present in the vocabulary of these neural embeddings. \textbf{Setup 2} is generated from a corpus in a different clinical subdomain, and also exclusively contains in-vector-vocabulary corrections. \textbf{Setup 3} presents the most difficult scenario, where we use the same corpus as for Setup 2, but only include corrections which are not present in the embedding vocabulary (OOV). In other words, here our model has to deal with both domain change and data sparsity. 

Correcting OOV tokens in Setup 3 is made possible by using a combination of word and character n-gram embeddings. We train these embeddings with the fastText model \cite{bojanowski}, an extension of the popular Word2Vec model \cite{mikolov}, which creates vector representations for character n-grams and sums these with word unigram vectors to create the final word vectors. FastText allows for creating vector representations for misspelling replacement candidates absent from the trained embedding space, by only summing the vectors of the character n-grams.

\begin{figure}
\includegraphics[width=15cm,resolution=300,keepaspectratio=False]{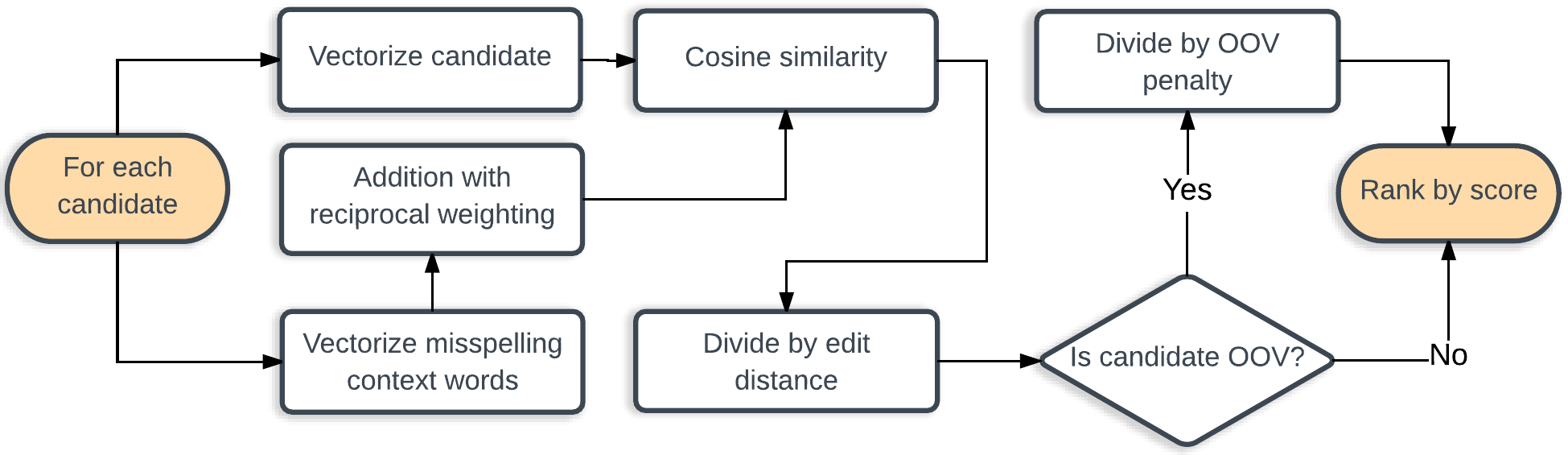}
\caption{\label{flowchart}The final architecture of our model. Within a specified window size (9 for English, 10 for Dutch), it vectorizes every context word on each side if it is present in the vector vocabulary, applies reciprocal weighting, and sums the representations. It then calculates the cosine similarity with each candidate vector, and divides this score by the Damerau-Levenshtein edit distance between the candidate and misspelling. If the candidate is OOV, the score is divided by an OOV penalty.}
\end{figure}

We report our development experiments with the different setups in section \ref{dev}. The final architecture of our model for both English and Dutch is described in Figure \ref{flowchart}. We evaluate this model on our test data in section \ref{test}.

\section{Materials}

We tokenize all English data with the Pattern tokenizer \cite{pattern}, and all Dutch data with Ucto\footnote{\url{https://languagemachines.github.io/ucto/}}. All text is lowercased\footnote{While this has consequences for the nature of the task, it is a salient aspect of training good embeddings. Lowercasing reduces sparsity, therefore leading to more reliable representations, especially in the case of low frequency words.}, and we remove all tokens that include anything different from alphabetic characters or hyphens. Table \ref{tab:data} gives a comprehensive overview of the English and Dutch development and test corpora we describe in section \ref{devcorp} and \ref{testcorp}.

\begin{table}
\caption{\label{tab:data}A comprehensive overview of our corpora described in section \ref{devcorp} and \ref{testcorp}.}
\centering
\begin{tabular}{|m{5em} | c | c | c | c | }
\hline
\textbf{Language} & \textbf{Corpus type} & \textbf{Domain} & \textbf{Data used} & \textbf{Instances} \\
\hline
& \textsc{Development: Setup} 1 & critical care & \textsc{mimic-iii} & \textsc{5,000}\\
\cline{2-5}
\multirow{2}{*}{\textbf{\textsc{English}}} & \textsc{Development: Setup 2} & breast/colon cancer & \textsc{thyme} & \textsc{5,000}\\
\cline{2-5}
& \textsc{Development: Setup 3} & breast/colon cancer & \textsc{thyme}  & \textsc{1,500}\\
\cline{2-5}
& \textsc{Test} & critical care  & \textsc{mimic-iii}  & \textsc{873}\\
\midrule
& \textsc{Development: Setup 1} & critical care & \textsc{uza} & \textsc{5,000}\\
\cline{2-5}
\multirow{2}{*}{\textbf{\textsc{Dutch}}} & \textsc{Development: Setup 2} & breast/colon cancer & \textsc{uza} & \textsc{5,000}\\
\cline{2-5}
& \textsc{Development: Setup 3} & breast/colon cancer & \textsc{uza}  & \textsc{350}\\
\cline{2-5}
& \textsc{Test} & miscellaneous  & \textsc{uza} & \textsc{490}\\
\hline
\end{tabular}
\end{table}

\subsection{Lexicons} \label{lexicon}

To construct reference lexicons, we fuse general dictionaries with specialized resources. For our English lexicon, we use a union of the general dictionary from Jazzy\footnote{\url{http://jazzy.sourceforge.net}}, a Java open source spell checker (47,160 items), and the UMLS\textsuperscript{\textregistered} SPECIALIST lexicon\footnote{\url{https://lexsrv3.nlm.nih.gov/LexSysGroup/Projects/lexicon/current/web/index.html}} (304,840 items), which contains a broad range of specialized clinical terms. This amounts to 319,579 unique lexical items. For our Dutch lexicon, we use as general dictionary the publicly available word list from Stichting OpenTaal\footnote{\url{https://www.opentaal.org}} (320,913 tokens), which has the official quality label of the Dutch Language Union. As specialized resource, we extract terminology from two clinical resources, namely, the Belgian Bilingual Biclassified Thesaurus (23,794 items) constructed by the universities of Ghent and Brussels, and the UMLS\textsuperscript{\textregistered} Metathesaurus\footnote{\url{https://www.nlm.nih.gov/research/umls/knowledge_sources/metathesaurus/}} (77,646 items). This amounts to 371,559 unique lexical items.

\subsection{Neural embeddings} \label{embeddings}

We train a fastText skipgram model using the default parameters, except for the dimensionality, which we raise to 300, since we want to make sure that the embeddings are able to capture subtle semantic relationships in a training corpus of our size. For our English experiments, we train on 425M words from the MIMIC-III corpus, which contains medical records from critical care units. For our Dutch experiments, we train on 720M words from clinical records collected at the Antwerp University Hospital (UZA). These records span a decade in time, and cover various genres (notes, letters, protocols, reports) as well as a wide range of clinical subdomains, including gastroenterology, pulmonology, and critical care.

\begin{table}
\caption{\label{tab:engdevexamples}Examples of automatically generated spelling errors and some replacement candidates for the English development setups.}
\centering
\begin{tabular}{| c | c | c |}
\hline
& \textbf{Misspelling} & \textbf{Candidates} \\
\hline
\textbf{Setup 1} & \textit{unchanged} $\rightarrow$ \textit{unchainged}  & unchanged, unchained, uncharged, unhinged\\
\hline
\textbf{Setup 2} & \textit{chronic} $\rightarrow$ \textit{chornic} & chronic, choreic, cornice, chloric\\
\hline
\textbf{Setup 3} & \textit{accrued} $\rightarrow$ \textit{accued} & accrued, accused, accuse, accede\\
\hline
\end{tabular}
\end{table}

\begin{table}
\caption{\label{tab:nldevexamples}Examples of automatically generated spelling errors and some replacement candidates for the Dutch development setups.}
\centering
\begin{tabular}{| c | c | c | c | c |}
\hline
& \textbf{Misspelling} & \textbf{Candidates} \\
\hline
\textbf{Setup 1} & \textit{mediane} $\rightarrow$ \textit{medciane}  & mediane, mediale, medianen, Mediene\\
\hline
\textbf{Setup 2} & \textit{beperkt} $\rightarrow$ \textit{beprekt} & beperkt, betrekt, verrekt, gerekt, bevlekt\\
\hline
\textbf{Setup 3} & \textit{megacyste} $\rightarrow$ \textit{megacyte} & megacyste, megabyte, megabytes\\
\hline
\end{tabular}
\end{table}

\subsection{Development corpora} \label{devcorp}

In order to tune our model parameters in an unsupervised way, we automatically create self-induced error corpora. We generate these development corpora by randomly sampling lines from a reference corpus, randomly sampling a single word per line if the word is present in our reference lexicon, transforming these words with either 1 (80\%) or 2 (20\%) random Damerau-Levenshtein operations to a non-word, and then extracting these misspelling instances with a context window of up to 10 tokens on each side. Table \ref{tab:data} gives an overview of all the development corpora and the data used to generate them. Table \ref{tab:engdevexamples} and \ref{tab:nldevexamples} give examples from all development corpora for both languages. For \textbf{Setup 1}, we perform our corpus creation procedure for critical care records, a domain which is present in the data used to train our neural embeddings. We exclusively sample words present in our vector vocabulary, resulting in 5,000 tokens for both English and Dutch. For \textbf{Setup 2}, we perform our procedure for records which exclusively cover the domain of brain and colon cancer, which is not represented in our neural embedding corpora. For English, we use the THYME \cite{thyme} corpus. For Dutch, we use data which originally belonged to our neural embeddings training data, but which was located and held out before our experiments. We once again exclusively sample in-vector-vocabulary words, resulting in 5,000 tokens for both English and Dutch. For \textbf{Setup 3}, we again perform our procedure for the cancer corpora, but this time we exclusively sample OOV words, resulting in 1,500 tokens for English and 350 for Dutch. While this last setup can seem exaggerated or overly artificial, we want to explicitly isolate these cases from the other setups, since the distribution of OOVs is entirely  dependent on the vocabulary overlap between the data being corrected and the data used to train the neural embeddings. In other words, it is relative with respect to the specific use case of our model in practice. On the one hand, we use this setup to estimate how well our trained model can generalize to other subdomains and corpora with only partially overlapping vocabulary; on the other hand, we use this setup to regulate the role of OOV correction candidates, as we discuss in section \ref{dev}.

\subsection{Test corpora} \label{testcorp}

No benchmark test sets are publicly available for clinical spelling correction. A straightforward annotation task is costly and can lead to small corpora, such as the one by Lai et al., which contains just 78 misspelling instances. Therefore, we adopt a more cost-effective annotation approach. In a corpus, we spot misspellings by looking at items with a frequency of 5 or lower which are absent from our lexicon.\footnote{While this excludes frequent error types, and is therefore far from an optimal strategy, it is hard to estimate the possible deceiving effect of this strategy without knowing the frequency distribution of spelling errors in the MIMIC-III corpus.} We then extract and annotate instances of these misspellings along with their context. For English, we use the MIMIC-III data, resulting in 873 contextually different tokens of 357 unique error types.\footnote{A script to extract this data can be found at \url{https://github.com/clips/clinspell}.} For Dutch, we use a recent set of clinical records from the Antwerp University Hospital, which covers the same genres and domains as the neural embeddings training data. This results in 490 contextually different tokens of 359 unique error types. Tables \ref{tab:engtestexamples} and \ref{tab:nltestexamples} give examples from both test corpora.

\begin{table}
\caption{\label{tab:engtestexamples}Examples of empirically observed misspellings and some replacement candidates from our English test set, per Damerau-Levenshtein edit distance.}
\centering
\begin{tabular}{| c | c | c | c | c |}
\hline
& \textbf{Misspelling} & \textbf{Candidates} \\
\hline
\textbf{Edit distance 1} &  \textit{sclerosing}  $\rightarrow$ \textit{sclerosin} & sclerosing, sclerosis, sclerotin, sclerostin\\
\hline
\textbf{Edit distance 2} & \textit{symptoms} $\rightarrow$ \textit{sympots} & symptoms, symptom, spots, symbols\\
\hline
\textbf{Edit distance 3} & \textit{phlebitis} $\rightarrow$ \textit{phebilitis} & phlebitis, cheilitis, pyelitis, phallitis\\
\hline
\end{tabular}
\end{table}

\begin{table}
\caption{\label{tab:nltestexamples}Examples of empirically observed misspellings and some replacement candidates from our Dutch test set, per Damerau-Levenshtein edit distance.}
\centering
\begin{tabular}{| c | c | c | c | c |}
\hline
& \textbf{Misspelling} & \textbf{Candidates} \\
\hline
\textbf{Edit distance 1} & \textit{letsels}  $\rightarrow$ \textit{letels} & letsels, lepels, netels, zetels, zetsels\\
\hline
\textbf{Edit distance 2} & \textit{weinig} $\rightarrow$ \textit{wijnig} & weinig, pijnig, wijzig, tijdig, wijn\\
\hline
\textbf{Edit distance 3} & \textit{verminderde} $\rightarrow$ \textit{verminderderde} & verminderde, verminderende\\
\hline
\end{tabular}
\end{table}

\section{Results}

We first develop our model for each language by tuning the parameters with the development corpora. We then test this tuned model on the test data. We discuss the results and their implications in the next section. To evaluate the performance of our model, we use \textbf{first-best} accuracy as criterion, i.e., the percentage of misspellings which are properly corrected by the first-ranked replacement suggestion of our model. We use two variations of first-best accuracy, the terminology of which we borrow from Reynaert (2008): \textbf{true} first-best accuracy, which is the accuracy given the system's dictionary; and \textbf{upper-bound} first-best accuracy, which removes the effect of dictionary shortcomings, by adding all correct word forms for the errors to be corrected to the system's spelling dictionary. The latter criterion allows for measuring the upper bound on correction attainable by our system.

\subsection{Development} \label{dev}

To develop our model, we investigate a variety of parameters:

\begin{enumerate}[leftmargin=*, label=]
	\begin{multicols}{2}
	\item{\textbf{Vector composition functions}}
	\begin{enumerate}
		\item{addition}
		\item{multiplication}
		\item{max embedding by Wu et al. (2015)}
	\end{enumerate}
	\columnbreak
	\item{\textbf{Edit distance penalty}}
	\begin{enumerate}
		\item{Damerau-Levenshtein}
		\item{Double Metaphone}
		\item{Damerau-Levenshtein + Double Metaphone}
		\item{Spell score by Lai et al.}
	\end{enumerate}
	\end{multicols}
	\item{\textbf{Context parameters}}
	\begin{enumerate}
		\item{Window size (1 to 10)}
		\item{Reciprocal weighting}
		\item{Removing stop words using the English stop word list from scikit-learn \cite{pedregosa} or the Dutch stop word list from Pattern \cite{pattern}}
		\item{Including a vectorized representation of the misspelling}
	\end{enumerate}
\end{enumerate}

\noindent
We perform a grid search for Setup 1 and Setup 2 to discover which parameter combination leads to the highest accuracy averaged over both corpora. In this setting, we only allow for candidates which are present in the vector vocabulary. We then introduce OOV candidates for Setup 1, 2 and 3, and experiment with penalizing them, since their representations are less reliable. As these representations are only composed out of character n-gram vectors, with no word unigram vector, they are susceptible to noise caused by the particular nature of the n-grams; namely, sometimes the semantic similarity of OOV vectors to other vectors can be inflated in cases of strong orthographic overlap. OOV replacement candidates are more often redundant than necessary, as in most practical use cases of the correction model (where there is considerable vocabulary overlap between the embedding domain and the correction domain), the majority of correct misspelling replacements will be present in the trained vector space. Therefore, we try to penalize OOV representations to the extent that they do not cause noise in cases where they are redundant, but still rank first in cases where they are the correct replacement. We tune this OOV penalty by maximizing the accuracy for Setup 3 while minimizing the performance drop for Setup 1 and 2, using a weighted average of their correction accuracies. 

The final architecture of our model for both English and Dutch is described in full in Figure \ref{flowchart}, showing all used parameters. As the description shows, the models for both languages only differ in optimal window size (9 for English, 10 for Dutch). To compare our method against a reference noisy channel model in the most direct way, we implement the ranking component of Lai et al.'s model in our pipeline (\textbf{Noisy Channel}). This component requires corpus frequencies, which we extract from the same data that we use to train the embeddings. Our context-sensitive model (\textbf{Context}) outperforms the noisy channel for each corpus in our development phase, for both English and Dutch, as shown in Table \ref{tab:engdev} and \ref{tab:nldev}. Moreover, as the results for Setup 3 show, our method generalizes considerably better to OOV misspellings, as we explicitly intended in the development of our model.

\subsection{Test} \label{test}

Table \ref{tab:engtest} shows the English correction accuracies for \textbf{Context} and \textbf{Noisy Channel} as off-the-shelf tools, compared to two existing tools. The first tool is HunSpell, a popular open source spell checker used by Google Chrome and Firefox. The second tool is the original implementation of Lai et al.'s model, which they shared with us. Table \ref{tab:nltest} shows the Dutch correction accuracies for \textbf{Context} and \textbf{Noisy Channel} as off-the-shelf tools, as compared to HunSpell.

\begin{table}
\caption{\label{tab:engdev}True first-best correction accuracies for our 3 English development setups.}
\centering
\begin{tabular}{| c | c | c | c | }
\hline
& \textbf{Setup 1} & \textbf{Setup 2}& \textbf{Setup 3} \\
\hline
\textbf{Context} & 90.24 & 88.20 & 57.00 \\
\hline
\textbf{Noisy Channel} & 85.02 & 85.86 &  39.73\\
\hline
\end{tabular}
\end{table}

\begin{table}
\caption{\label{tab:nldev}True first-best correction accuracies for our 3 Dutch development setups.}
\centering
\begin{tabular}{| c | c | c | c | }
\hline
& \textbf{Setup 1} & \textbf{Setup 2}& \textbf{Setup 3} \\
\hline
\textbf{Context} & 87.94 & 89.10 & 82.00 \\
\hline
\textbf{Noisy Channel} & 86.90 & 85.80 &  66.57\\
\hline
\end{tabular}
\end{table}

The performance of our models on the test sets is held back by the incomplete coverage of our reference lexicons. For English, missing corrections are mostly highly specialized medical terms, or inflections of more common terminology. For Dutch, this includes relatively infrequent compounds as well. Compounds in Dutch, as opposed to English, are mostly orthographically concatenated into one lexical item. Since Dutch language users tend to be very productive with compounding, this leads to a whole range of standard language that is hard to cover exhaustively in a lexicon. We use the upper-bound first-best correction accuracy to examine the performance of our ranking models with disregard to such circumstances. Tables \ref{tab:engtest} and \ref{tab:nltest} show that the performance according to this metric is comparable to the true first-best correction accuracy for the development corpora.

\begin{table}
\caption{\label{tab:engtest}The correction accuracies for our English test experiments, evaluated for two different scenarios. \textit{True first-best accuracy}: gives the first-best accuracies of all off-the-shelf tools. \textit{Upper-bound first-best accuracy}: gives the first-best accuracies of our implemented models for the scenario where correct replacements missing from the lexicon are included in the lexicon before the experiment.}
\centering
\begin{tabular}{| c | c | c | c | c | }
\hline
\textbf{Evaluation} & \textbf{HunSpell} & \textbf{Lai et al.} & \textbf{Context} & \textbf{Noisy Channel} \\
\hline
\textsc{true first-best accuracy} & 52.69 & 61.97 & 88.21 & 87.85\\
\hline
\textsc{upper-bound first-best accuracy} &  &  & 93.02 & 92.66 \\
\hline
\end{tabular}
\end{table}

\begin{table}
\caption{\label{tab:nltest}The correction accuracies for our Dutch test experiments, evaluated for two different scenarios. \textit{True first-best accuracy}: gives the accuracies of all off-the-shelf tools. \textit{Upper-bound first-best accuracy}:  gives the accuracies of our implemented models for the scenario where correct replacements missing from the lexicon are included in the lexicon before the experiment.}
\centering
\begin{tabular}{| c | c | c | c | c | }
\hline
\textbf{Evaluation} & \textbf{HunSpell} & \textbf{Context} & \textbf{Noisy Channel} \\
\hline
\textsc{true first-best accuracy} & 64.29 & 76.53 & 79.71\\
\hline
\textsc{upper-bound first-best accuracy} &  & 87.75 & 92.45 \\
\hline
\end{tabular}
\end{table}

\section{Discussion} \label{discussion}

In terms of correction accuracy, our context-sensitive model and our own implementation of Lai et al.'s ranking model outperform off-the-shelf tools for both English and Dutch, establishing a state-of-the-art for spelling correction of clinical free-text. The salient difference in performance between Lai et al.'s system and our specific implementation of their noisy channel model highlights the influence of (lack of) training resources and development decisions on the general applicability of spelling correction models. Moreover, it shows the strength of the noisy channel model in scenarios where the scale of the resources is sufficient (in this case, 425M words for English and 720M words for Dutch) to reliably estimate prior probabilities from corpus frequencies.

However, sufficient empirical resources to estimate a fine-grained likelihood (namely, a large corpus of empirically observed errors from which a reliable error model can be extracted) are still absent for the clinical domain. Therefore, the likelihood of Lai et al.'s ranking model is estimated with a rudimentary spell score, which is a weighted combination of Damerau-Levenshtein and Double Metaphone edit distance. While this error model leads to a noisy channel model which is robust in performance, as shown by our test results, it also leads to a pragmatic performance ceiling where more heavily distorted replacement candidates are downplayed to safeguard robustness of performance, regardless of their possible empirical association with the misspelling. As a result, our noisy channel model is still prone to cases of frequency bias, including the example of frequency bias which we have provided in the introduction of this paper: our noisy channel model does not succeed in correcting the MIMIC-III misspelling \textit{goint} to the correct form \textit{going} due to the higher corpus frequency of, and therefore higher prior probability assigned to, the word type \textit{point}. While the difference in frequency is salient, it is not insurmountable for a likelihood reflecting a proper error model, which in this case would typically reflect that \textit{goint} is more probable to be a typo of \textit{going} than of \textit{point}. However, the rudimentary spell score does not reflect that notion. This example illustrates that, regardless of the theoretical validity of the noisy channel, we are still very much bound to the practical reality of its implementation, including the state of resources.

\begin{figure}
\begin{center}
\includegraphics[width=12cm,resolution=300,keepaspectratio=False]{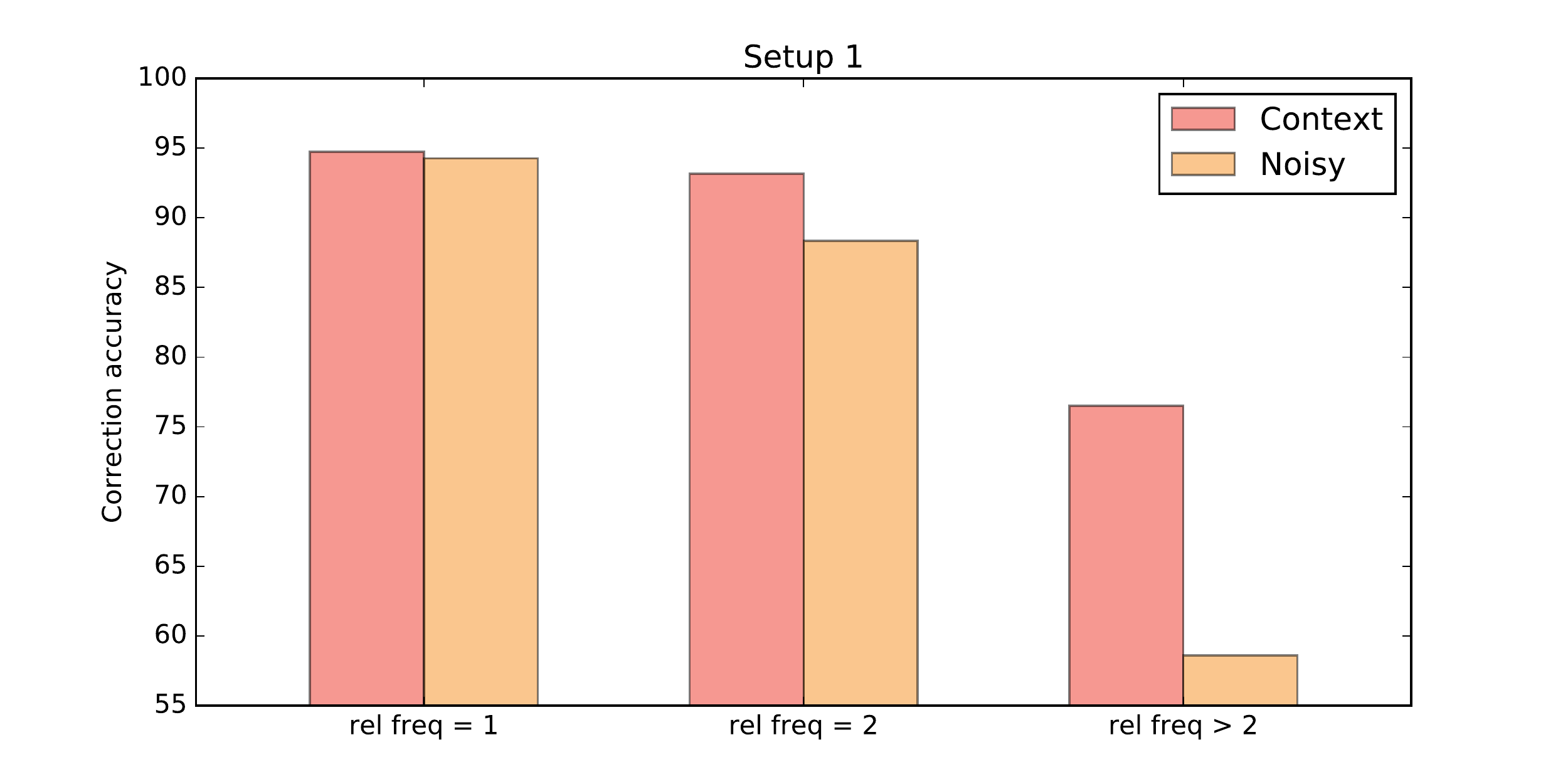}
\includegraphics[width=12cm,resolution=300,keepaspectratio=False]{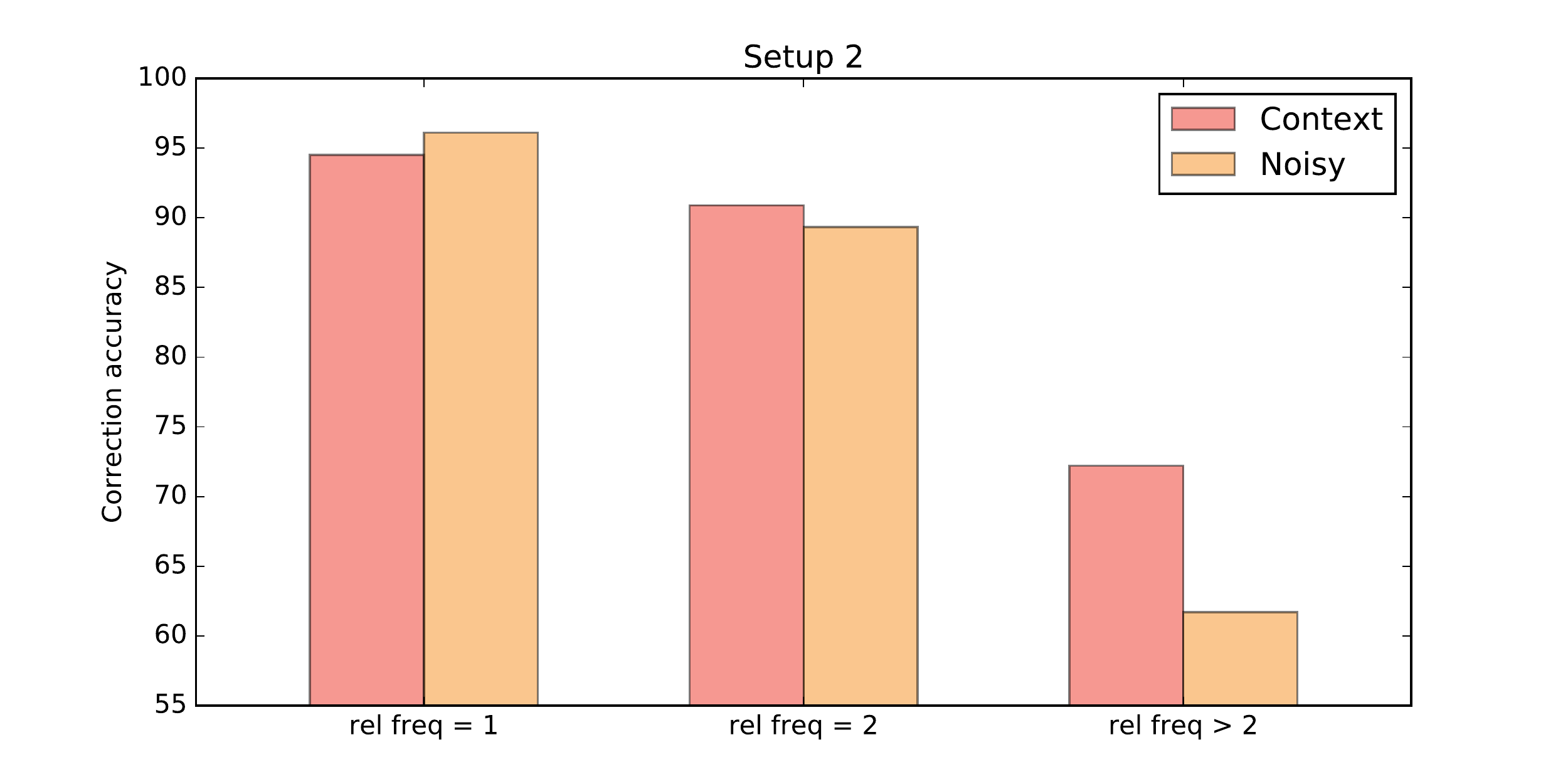}
\includegraphics[width=12cm,resolution=300,keepaspectratio=False]{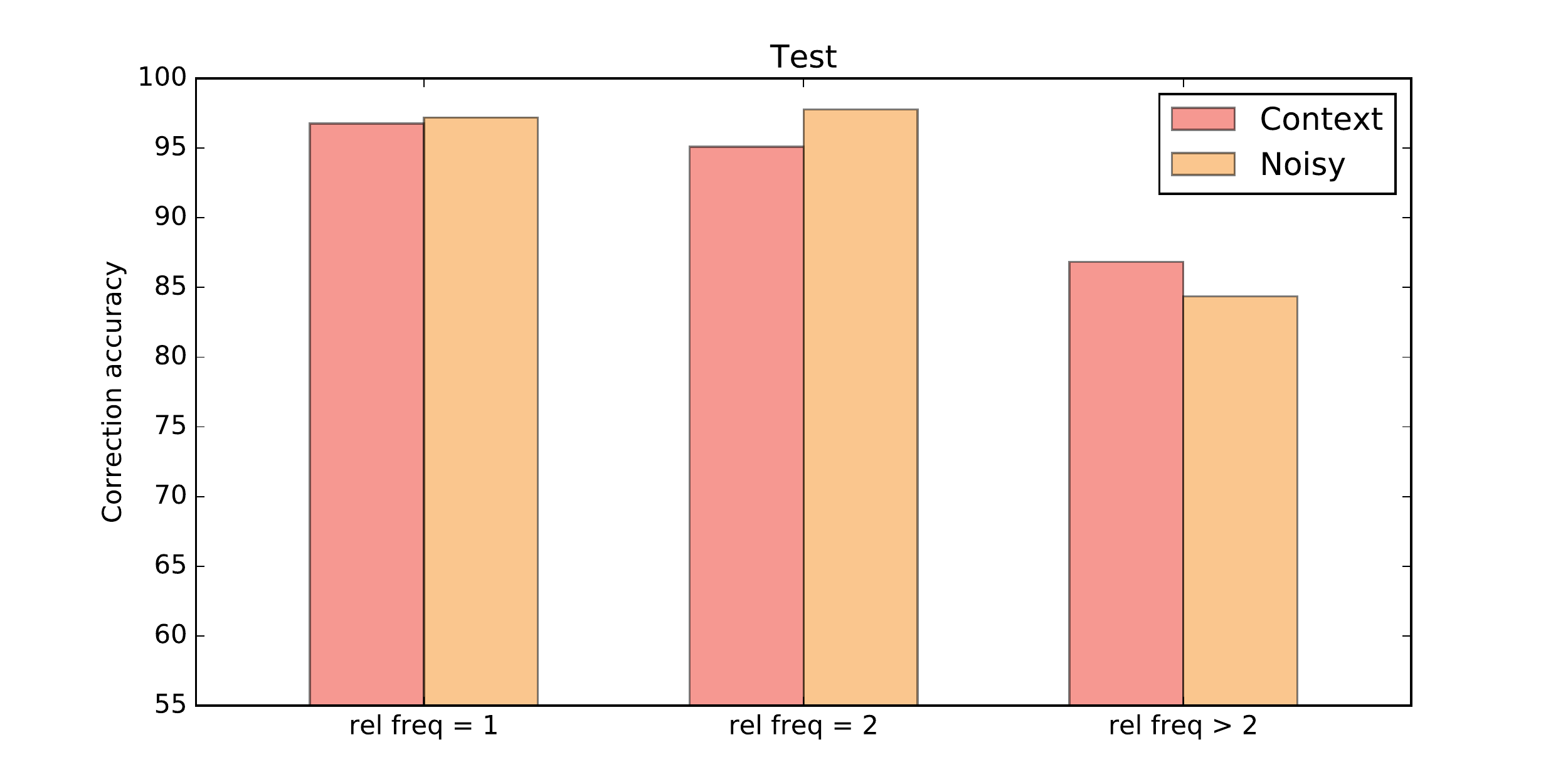}
\caption{\label{frequency_graph_eng}The English correction accuracies for \textbf{Context} and \textbf{Noisy Channel} for Setup 1, Setup 2, and the test set, grouped per relative frequency of the correct replacement compared to other replacement candidates. \textit{rel freq = 1}: highest corpus frequency of all candidates. \textit{rel freq = 2}: second highest corpus frequency of all candidates. \textit{rel freq $>$ 2}: corpus frequency lower than second highest of all candidates.}
\end{center}
\end{figure}

\begin{figure}
\begin{center}
\includegraphics[width=12cm,resolution=300,keepaspectratio=False]{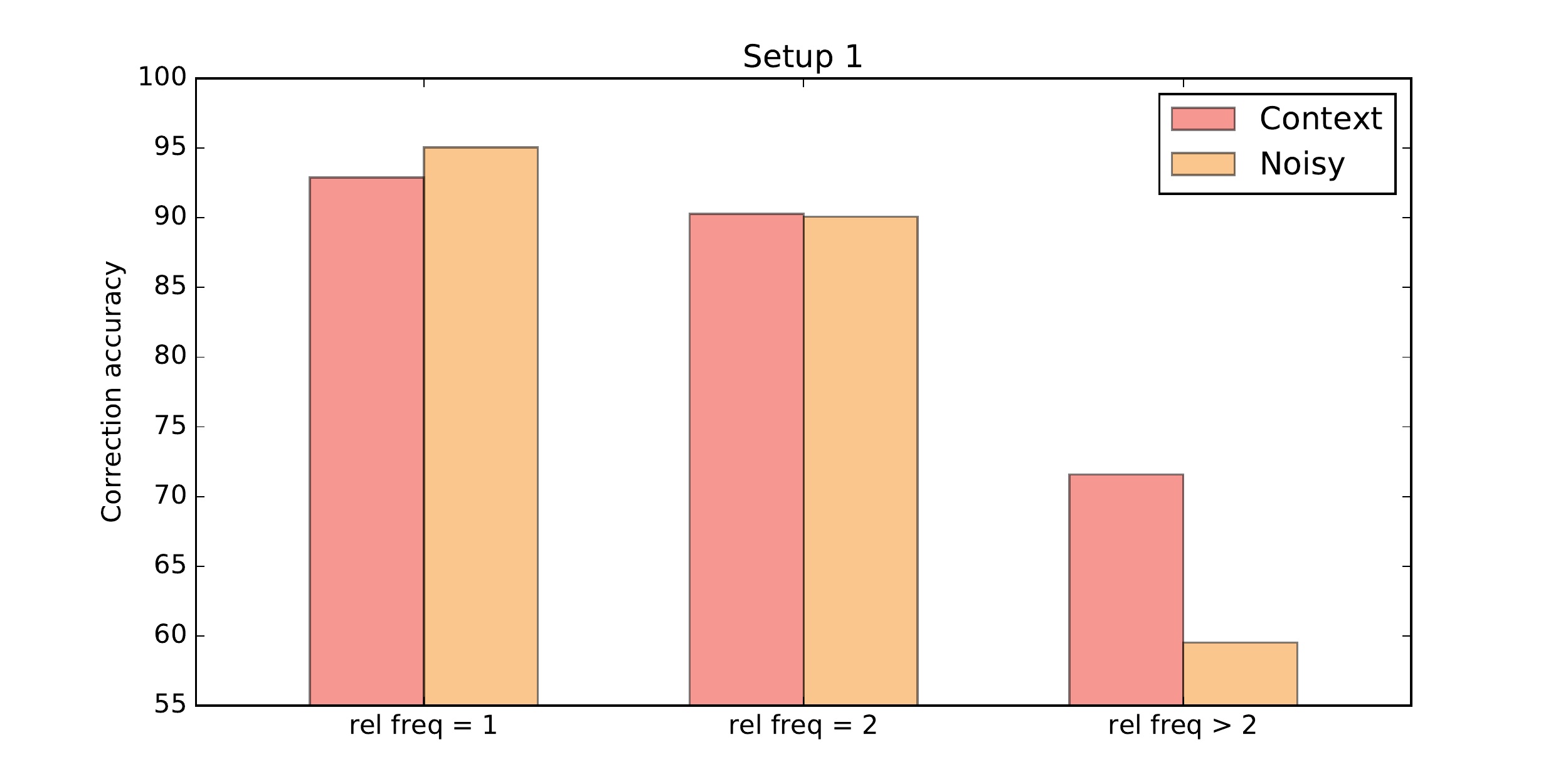}
\includegraphics[width=12cm,resolution=300,keepaspectratio=False]{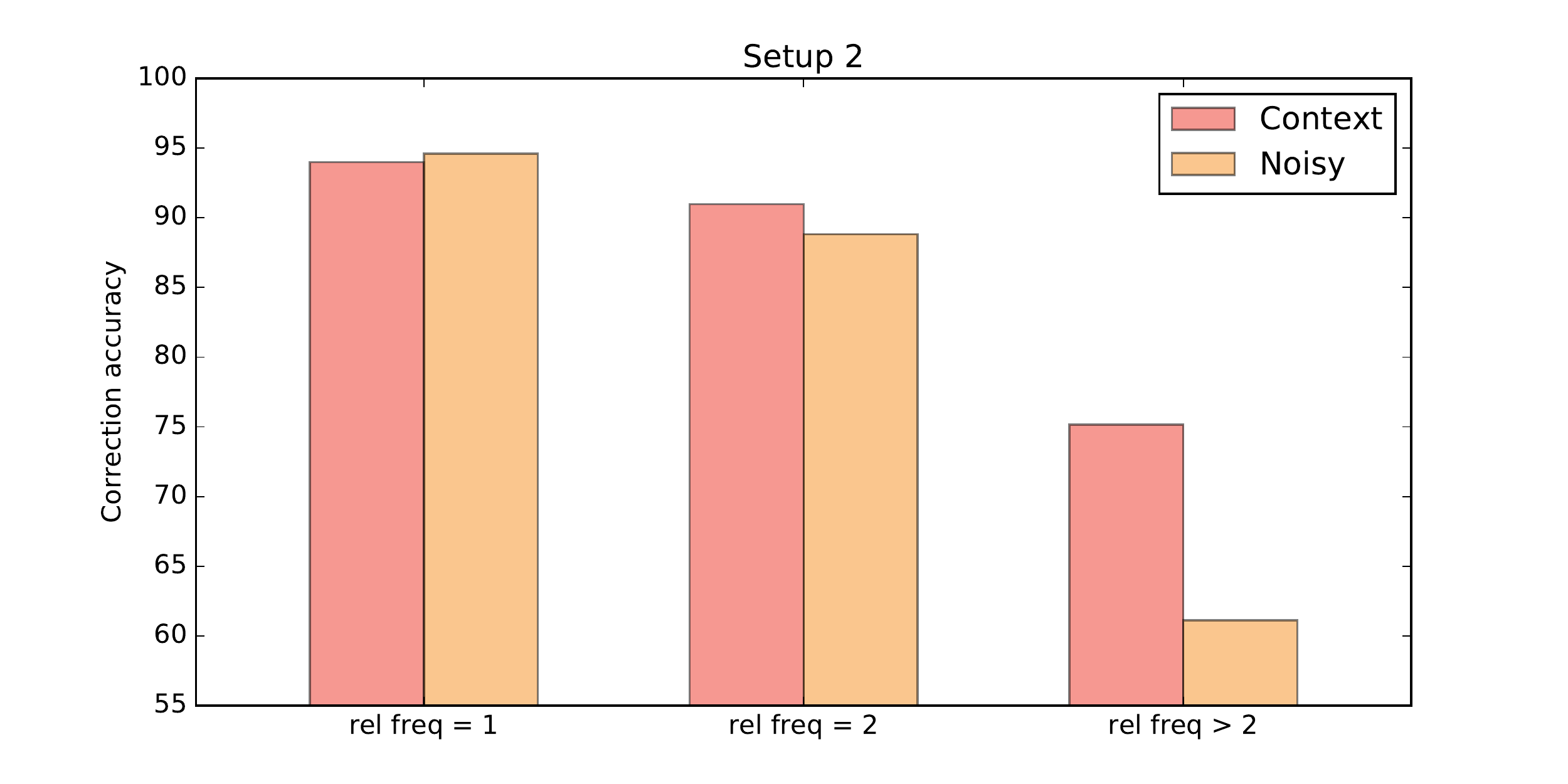}
\includegraphics[width=12cm,resolution=300,keepaspectratio=False]{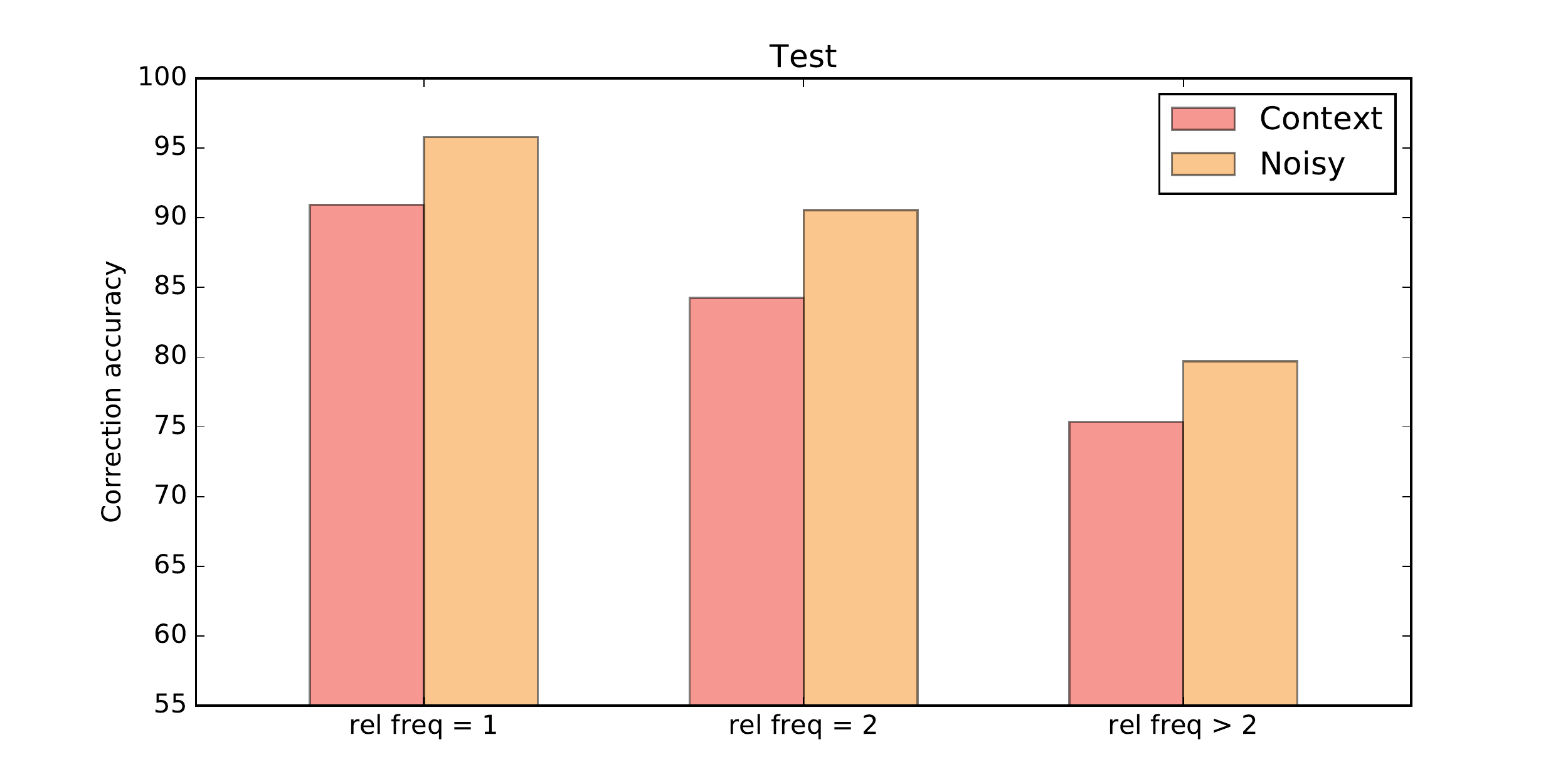}
\caption{\label{frequency_graph_nl}The Dutch correction accuracies for \textbf{Context} and \textbf{Noisy Channel} for Setup 1, Setup 2, and the test set, grouped per relative frequency of the correct replacement compared to other replacement candidates. \textit{rel freq = 1}: highest corpus frequency of all candidates. \textit{rel freq = 2}: second highest corpus frequency of all candidates. \textit{rel freq $>$ 2}: corpus frequency lower than second highest of all candidates.}
\end{center}
\end{figure}

\begin{figure}
\begin{center}
\includegraphics[width=9cm,resolution=300,keepaspectratio=False]{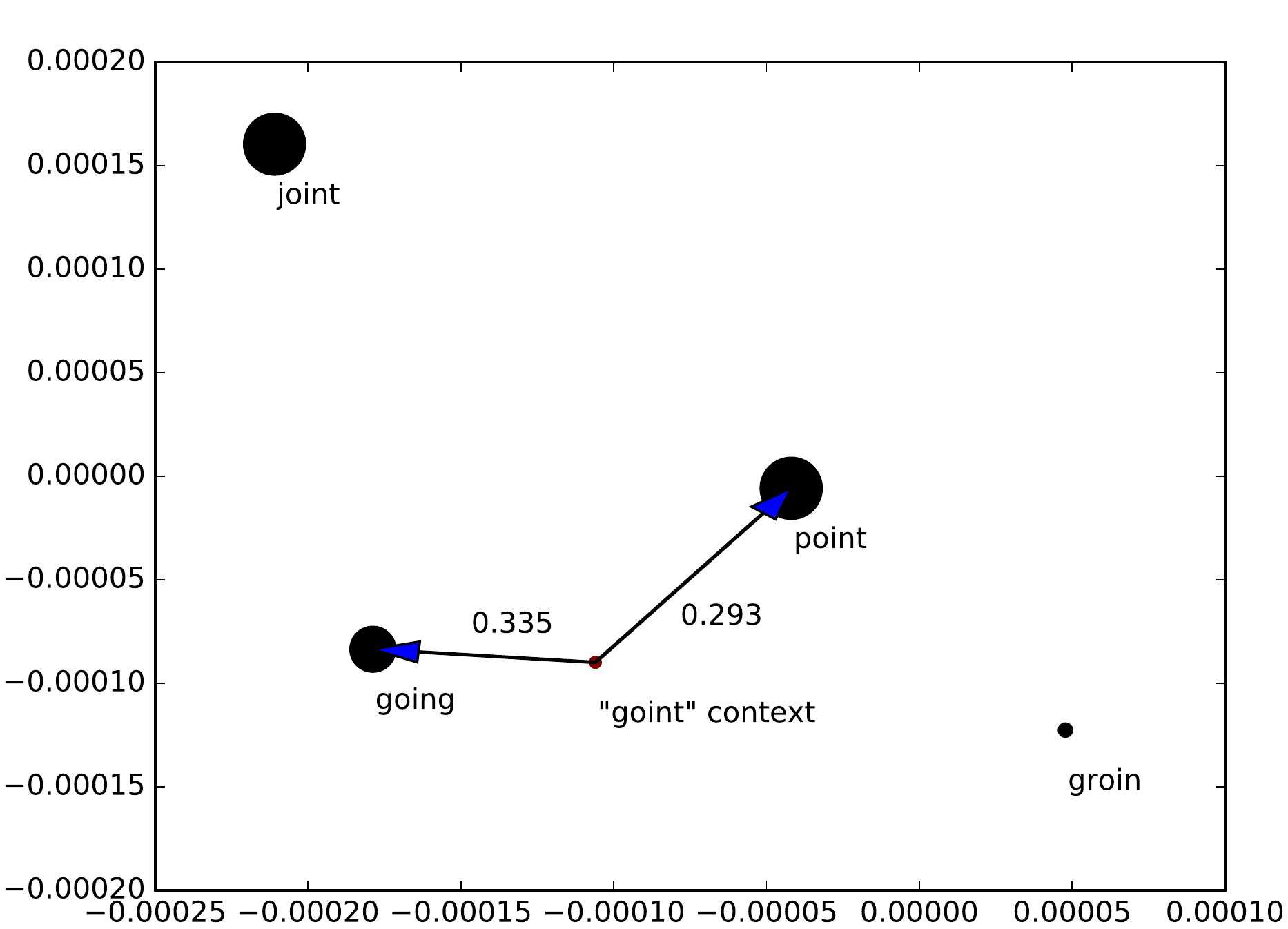}
\caption{\label{embedding_space}2-dimensional t-SNE projection of the vectorized context of the English test misspelling \textit{goint}  and 4 replacement candidates in the trained MIMIC-III vector space. Dot size denotes corpus frequency, numbers denote cosine similarity. The English misspelling context is \textit{new central line lower extremity bypass with sob now [goint] to [be] intubated}. While the noisy channel chooses the more frequent \textit{point}, our model correctly chooses the most semantically fitting \textit{going}.}
\end{center}
\end{figure}

Our method tries to improve on the clinical spelling correction process considering the availability of actual incomplete resources. As it stands, a noisy channel model like the one by Lai et al. still occasionally suffers from frequency bias; it is not able to correct a specific misspelling type to different corrections in different contexts, and is not sufficiently equipped to deal with word types that are not observed in training data. Our unsupervised context-sensitive model targets these weaknesses. Figures \ref{frequency_graph_eng} and \ref{frequency_graph_nl} show the correction accuracies for three scenarios: one where the most frequent candidate is the correct one (\textit{rel freq = 1}), one where the second most frequent candidate is the correct one (\textit{rel freq = 2}), and one where the correct candidate has a lower relative frequency (\textit{rel freq > 2}). Figure \ref{frequency_graph_eng} confirms the hypothesis that our context-sensitive model counters the frequency bias of a noisy channel model for our English experiments. The results for our development corpora show that in cases where \textit{rel freq = 1}, the noisy channel scores similar or slightly better, as expected. This trend is reflected in the test results. In cases where \textit{rel freq = 2}, our model scores slightly better. This trend is not reflected in the test results. In fact, it is reversed. Lastly, in cases where \textit{rel freq > 2}, our model scores much better. This trend is reflected in the test results, if to a smaller extent. However, the relatively small sample size (a difference of 6 correct instances on a total of 243) should be kept in mind. Figure \ref{embedding_space} visualizes an example of frequency bias, where the \textit{goint} misspelling which we discussed earlier is correctly handled by our model as opposed to the noisy channel model.

Figure \ref{frequency_graph_nl} shows that the performance our context-sensitive model exhibits the same characteristics for the Dutch development corpora as for the English development corpora. However, this time none of the trends are reflected in the test results, which leads to our model being outperformed by the noisy channel model. This discrepancy raises the question to what extent the artificial nature of the development corpora leads to reliable models for empirical data. If the distributions of the several data types differ greatly, this undermines our unsupervised approach, which implicitly assumes that the distributions will not differ that greatly. To investigate this, we performed a grid search for both the English and Dutch test corpus, to examine which parameter combination leads to the best-performing model. For the English test data, this parameter combination is similar to our actual model derived from our development experiments. In other words, the underlying assumption of our unsupervised approach is confirmed. 

For the Dutch test data, however, the optimal parameter combination differs dramatically from our developed model. It includes two parameters which are absent from our developed model described in Figure \ref{flowchart}: the context representation also includes a vectorized representation of the misspelling itself, and the edit distance weighting adds Double Metaphone edit distance to the Damerau-Levenshtein edit distance. Moreover, the optimal context window size is 2, which is considerably smaller than for the originally developed model. With this parameter combination, the output of the model for the Dutch test data is exactly similar to the output of the noisy channel model. These analyses suggest that the distribution of the Dutch test data differs greatly from that of the development data. This discrepancy can be caused by the sparsity of the Dutch test data, which covers the same amount of error types as the English test data, but much fewer contextually different instances. The only conclusion we can draw is that the nature of our test set is possibly skewed in a way that does not allow for a thorough comparative evaluation of our models. As it stands, however, we have no empirical evidence that our Dutch context-sensitive model actually counters the frequency bias of our noisy channel. While we want to avoid too much speculation as to the reason why, these results invite inquiry into how important context actually is for Dutch clinical spelling correction.

When we look at the output of our context-sensitive model for both English and Dutch, we can categorize the errors it makes in 3 different types. The first type of errors concerns, predictably, misspellings for which the contextual clues are too unspecific. This lack of useful contextual information is sometimes caused by occurrences of other misspellings in the context window, and poses a fundamental challenge to our method. The second type of errors concerns cases where the contextual clues are actually misguiding. This happens for instance in cases where a word type has multiple senses which are not strongly related. Our Dutch test set contains the misspelling \textit{\sout{poslen} $\rightarrow$ polsen}, where from the context it appears that \textit{polsen} has the more infrequent sense of `polling someone about something' instead of the prevalent sense `wrists'. Since this word type shares one vector representation for both senses, the contextual information does not turn out to be strong enough for correcting the misspelling to the correct word type. Lastly, while our development experiments have tried to minimize the noise spread by OOV candidates, it is still noticeable in some instances.

\section{Conclusion and future research}

In this paper, we have proposed an unsupervised context-sensitive model for clinical spelling correction which uses word and character n-gram embeddings. This simple ranking model, which can be tuned to a specific language and domain by generating self-induced error corpora, tries to counter the frequency bias of a noisy channel model by exploiting contextual clues. 

As an implemented spelling correction tool for English clinical free-text, our method outperforms both a broadly used and a domain-specific off-the-shelf tool for empirically observed misspellings in MIMIC-III. Moreover, a detailed analysis of its performance shows that it does in fact counter the frequency bias of a noisy channel model. However, the relatively small sample size for this analysis should be kept in mind.

As an implemented spelling correction tool for Dutch clinical free-text, our method outperforms a broadly used off-the-shelf tool for empirically observed misspellings in collected data from the Antwerp University Hospital. However, our Dutch test set offers no empirical evidence that it counters the frequency bias of a noisy channel model. It is unclear whether this is caused by the sparsity of the test set.

Future research can investigate whether our method transfers well to other genres and domains. Secondly, it can address the three problem areas we have identified at the end of our discussion in section \ref{discussion}, namely, unspecific contextual clues, multiple word senses of a single word type, and noise spread by OOV candidates. Lastly, it is worthwhile to investigate how successfully our model can be applied to real-word errors.

\section{Acknowledgements}

This research was carried out in the framework of the Accumulate VLAIO SBO project, funded by the government agency Flanders Innovation \& Entrepreneurship (VLAIO). We would also like to thank Kim Luyckx for providing access to the Dutch data; Elyne Scheurwegs for preparing and managing the Dutch data; St\'ephan Tulkens for his logistic support with coding; and Kenneth Lai, Maxim Topaz, Foster R. Goss, and Li Zhou for sharing their system with us.

\nocite{*} 

\bibliographystyle{clin} 
\bibliography{CLINtemplate}  

\end{document}